\documentclass[11pt]{article}

\usepackage[preprint]{acl}
\usepackage{times}
\usepackage{latexsym}
\usepackage[T1]{fontenc}
\usepackage[utf8]{inputenc}
\usepackage{microtype}
\usepackage{inconsolata}
\usepackage{graphicx}

\usepackage{amsmath,amssymb,booktabs,multirow}
\usepackage{subcaption}
\usepackage{threeparttable}
\usepackage{adjustbox}
\usepackage[table]{xcolor}
\usepackage{bbm}
\usepackage{tabularx}
\usepackage{pifont}
\usepackage{listings}
\usepackage{url}

\newcommand{\method}{\textsc{SubLAT}}

\newcommand{\yes}{\ding{51}}
\newcommand{\no}{\ding{55}}
\newcolumntype{C}{>{\centering\arraybackslash}X}
\lstdefinestyle{methodcode}{
  language=Python,
  basicstyle=\ttfamily\footnotesize,
  keywordstyle=\color{blue!65!black},
  commentstyle=\color{gray},
  stringstyle=\color{green!40!black},
  numbers=none,
  frame=tb,
  breaklines=true,
  columns=fullflexible,
  keepspaces=true,
  showstringspaces=false,
  tabsize=4
}
\newcommand{\pmval}[2]{$#1_{\pm#2}$}

\title{Latent Path Exploration via Potential-Guided Subtrajectory Balance}
\author{
  Xiaoling Zeng \\
  Worcester Polytechnic Institute \\
  \texttt{xzeng4@wpi.edu}
  \And
  Qi Zhang \\
  Worcester Polytechnic Institute \\
  \texttt{qzhang9@wpi.edu}
}

\begin{document}
\maketitle

\begin{abstract}
Continuous latent reasoning enables language models to perform intermediate computation without generating explicit reasoning traces. Advances in stochastic latent reasoning allow multiple paths to be sampled in continuous space, but stochasticity alone does not ensure effective exploration. A key challenge is guiding stochastic transitions between intermediate latent states toward useful reasoning paths. We introduce \method{}, which couples a relative answer potential with Subtrajectory Balance to learn stochastic latent exploration.
The relative potential evaluates the answer support of sampled states against replacement states, with a correctness signal at the terminal state. Subtrajectory Balance relates potential differences to forward and backward transition densities across subtrajectories of different lengths. The model is jointly trained with answer supervision, CoT supervision, and representation distillation. 
Experiments on arithmetic and deductive reasoning show increasing solution coverage with larger sampling budgets across in-domain and out-of-domain benchmarks, as well as improvements on several structural OOD benchmarks. Trajectory analyses reveal progressive divergence among sampled latent paths and higher mean relative potentials for correct trajectories across arithmetic benchmarks.
\end{abstract}

\section{Introduction}
Chain-of-Thought (CoT) reasoning enables Large Language Models (LLMs) to solve complex problems through explicit intermediate steps \citep{wei2022chain}. Continuous latent reasoning instead performs intermediate reasoning in continuous space without verbalizing each step. COCONUT recursively feeds hidden states back into the model to form continuous reasoning trajectories \citep{hao2024training}, and CODI distills explicit CoT into latent reasoning by aligning teacher and student representations before answer generation \citep{shen2025codi}. These methods allow explicit CoT to guide latent reasoning without requiring the student to reproduce the teacher's reasoning in text.

Beyond learning a single latent trajectory, exploring alternative reasoning paths can uncover solutions missed by a single attempt \citep{wang2023selfconsistency}. Recent work extends this capability to continuous latent space through inference-time perturbations and learned stochastic policies \citep{you2026parallel,wang2026gts,zhou2026lepo,you2026slpo}. However, perturbation-based exploration is sensitive to noise strength and may increase trajectory diversity without improving solution coverage. Learned stochastic policies avoid relying solely on external perturbations, but training their intermediate transitions remains challenging when supervision is concentrated on final outcomes \citep{williams2026prioritize, zhao2026ltc}.

One way to guide stochastic transitions is to evaluate intermediate states through their support for the target answer. Answer likelihood provides such a signal \citep{tao2026trace}, but also reflects information from the question context, making the contribution of individual latent states difficult to assess. We therefore compare each sampled state with a replacement state under the same question and target answer, drawing on the idea of counterfactual comparison \citep{meulemans2023cocoa}. This comparison measures relative answer support between latent states rather than absolute likelihood. 
\begin{figure*}[h]
    \centering
    \includegraphics[width=\textwidth]{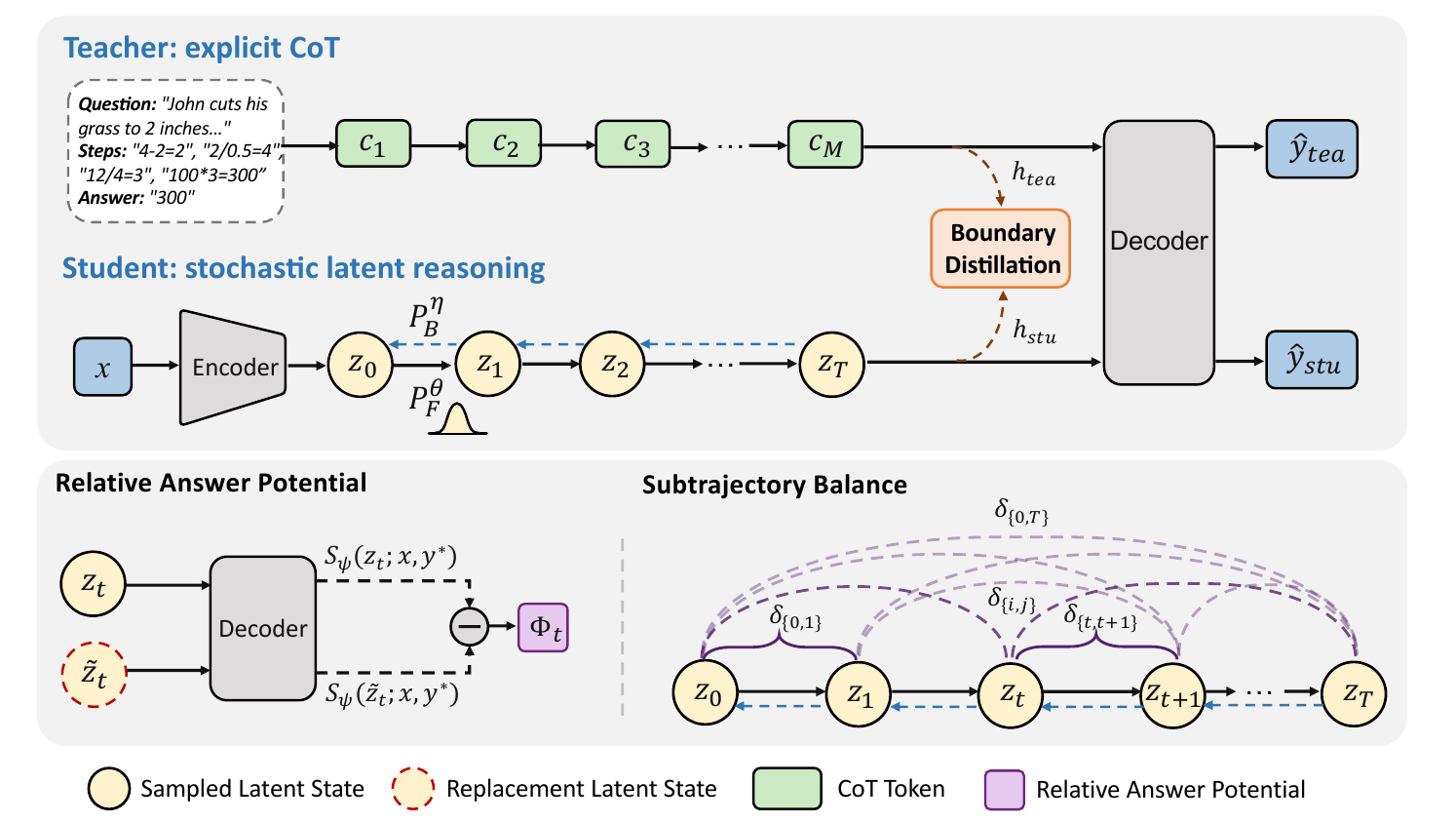}
    \caption{Illustration of \method{} for learning stochastic latent exploration through relative answer potentials and Subtrajectory Balance. The explicit CoT teacher provides a reference representation at the reasoning boundary, while the student samples latent trajectories through a stochastic forward policy $P_F^\theta$. The relative answer potential $\Phi_t$ provides intermediate credit for sampled latent states. Subtrajectory Balance couples potential differences with forward and backward transition flows for multi-scale credit assignment along latent trajectories.}
    \label{fig:method_overview}
\end{figure*}
To guide latent exploration, we propose \textbf{\method{}}, which couples Subtrajectory Balance with a relative answer potential to learn stochastic latent reasoning trajectories. A Gaussian forward policy samples successive latent states, while the balance objective relates potential differences to accumulated forward and backward transition log densities across subtrajectories of different lengths \citep{madan2023learning,lahlou2023theory}. This objective incorporates intermediate answer support and terminal correctness into stochastic policy learning. Following CODI \citep{shen2025codi}, we jointly train an explicit CoT pathway and distill its representations into the latent student, alongside answer supervision. At inference, \method{} requires only the learned forward policy and answer decoder, without access to the teacher.
Experiments on arithmetic and deductive reasoning show increasing solution coverage with larger sampling budgets across in-domain and OOD benchmarks. Trajectory analyses further reveal progressive diversification of sampled latent paths across reasoning steps.

Our main contributions are as follows:
\begin{itemize}
\item We develop a relative answer potential that measures the answer support of intermediate latent states through controlled comparisons.
\item We formulate potential-guided stochastic policy learning using Subtrajectory Balance, linking state evaluations to transitions at multiple temporal scales.
\item Experiments show increasing solution coverage with latent sampling, while further analyses distinguish effective exploration from latent dispersion and relate answer potentials to solution correctness.
\end{itemize}

\section{Related Work}
\paragraph{Continuous latent reasoning and exploration.}
Early work on implicit reasoning showed that explicit CoT steps can be progressively internalized into hidden computation \citep{deng2023implicit}. Continuous latent reasoning further replaces explicit intermediate steps with sequences of continuous hidden representations. COCONUT recurrently feeds hidden states back into the language model and gradually replaces textual reasoning steps with continuous thoughts \citep{hao2024training}, while CODI transfers explicit CoT reasoning into continuous space through self-distillation \citep{shen2025codi}. LRT uses a lightweight reasoning network to generate latent reasoning embeddings from the language model's hidden representations, which are then fed back to the base model for answer generation \citep{jiang2026rethinking}. More recent work considers multiple latent reasoning paths. Parallel latent test-time scaling applies MC-dropout and Gaussian perturbations to sample alternative trajectories at inference time \citep{you2026parallel}. GTS trains a Gaussian thought sampler on a frozen backbone with GRPO-style optimization \citep{wang2026gts}, while SLPO learns a surrogate transition density for policy optimization over stochastic latent trajectories \citep{you2026slpo}. \method{} instead learns the stochastic transition policy jointly with the latent reasoner, using teacher supervision and state-level credit rather than inference-time perturbation.
\paragraph{Credit assignment for latent reasoning.}
Final-answer supervision provides little information about intermediate latent computation. RLTT propagates outcome rewards across the iterations of looped language models \citep{williams2026prioritize}, and LTC values a latent thought by sampling multiple answers from its downstream context \citep{zhao2026ltc}. TIPS uses correct-answer likelihood to define an information potential and assigns turn-level rewards from changes in this potential \citep{xie2026tips}. Our relative answer potential also uses answer likelihood, but evaluates a visited state against a matched replacement while keeping the question and target answer fixed. This contrast follows the idea of counterfactual credit assignment, where contribution is evaluated relative to an alternative rather than from outcome value alone \citep{meulemans2023cocoa}.
\paragraph{Flow-based trajectory learning.}
GFlowNets provide a framework for learning stochastic policies by relating probability flow to reward \citep{bengio2023gflownet}. Trajectory Balance enforces flow consistency over complete trajectories \citep{whitammer2022trajectory}, while Subtrajectory Balance extends this constraint to partial trajectories of different lengths \citep{madan2023learning,pan2023better}. Continuous GFlowNets formulate these balance conditions with transition densities in continuous state spaces \citep{lahlou2023theory}. Flow-based objectives have also been applied to reasoning. Flow of Reasoning learns distributions over explicit graph-structured reasoning paths \citep{pmlr-v267-yu25k}, while Latent Thought Flow applies Subtrajectory Balance to variable-length latent reasoning with answer-quality and computation-cost rewards and deterministic backward transitions \citep{zou2026ltf}. In contrast, \method{} combines a relative answer potential with a learned Gaussian backward policy to guide fixed-length stochastic latent trajectories.

\section{Method}
\label{sec:method}
We introduce \method{}, a framework for learning stochastic latent reasoning paths by coupling a relative answer potential with Subtrajectory Balance. CoT supervision and representation distillation guide latent reasoning, while the potential-based balance objective optimizes the stochastic transition policy.

\subsection{Stochastic Latent Reasoning}
\label{sec:latent-exploration}
Given a question $x$, \method{} generates a sequence of continuous latent states $z_{1:T}=(z_1,\ldots,z_T)$ before answer generation. A forward policy $P_F^\theta$ samples successive latent states, guided by an explicit CoT pathway with reference rationale $c$ and gold answer $y^\star=(y^\star_1,\ldots,y^\star_L)$. The teacher and student share language model parameters $\psi$, and $\theta$ denotes the forward policy parameters.
\paragraph{Stochastic latent paths.}
The question is encoded into a deterministic source state $z_0=f_\psi(x)$. From this state, the forward policy generates a latent trajectory $\tau=(z_1,\ldots,z_T)$ with density $P_F^\theta(\tau\mid x)=\prod_{t=0}^{T-1}P_F^\theta(z_{t+1}\mid z_t,x)$. Each transition follows a state-conditioned diagonal Gaussian with a residual mean update:
\begin{equation}
P_F^\theta(\cdot\mid z_t,x)
=
\mathcal{N}\!\left(z_t+v_\theta,\,
\operatorname{diag}\!\left[(\sigma_F^\theta)^2\right]\right),
\label{eq:forward-policy}
\end{equation}
where the residual update $v_\theta(z_t,x)$ determines the transition mean and $\sigma_F^\theta(z_t,x)>0$ controls its spread. To backpropagate through the sampled latent states, we use the reparameterization $z_{t+1}=z_t+v_\theta(z_t,x)+\sigma_F^\theta(z_t,x)\odot\epsilon_t$ with $\epsilon_t\sim\mathcal{N}(0,I_d)$. 
\paragraph{Student answer supervision.}
After $T$ latent transitions, the terminal state $z_T$ is used as a single latent input for answer generation. Given the gold answer $y^\star=(y^\star_1,\ldots,y^\star_L)$, we minimize
\begin{equation}
\mathcal{L}_{\mathrm{ans}}
=
-\frac{1}{L}
\sum_{s=1}^{L}
\log p_\psi
\left(
y^\star_s\mid x,z_T,y^\star_{<s}
\right).
\label{eq:student-loss}
\end{equation}

\paragraph{CoT supervision and distillation.}
Following CODI \citep{shen2025codi}, we jointly train an explicit CoT pathway and distill its representations into the latent student. Given a reference rationale $c$, the teacher models the concatenated rationale and answer $r=[c;y^\star]$:
\begin{equation}
\mathcal{L}_{\mathrm{CoT}}
=
-\frac{1}{N_r}
\sum_{s=1}^{N_r}
\log p_\psi
\left(
r_s\mid x,r_{<s}
\right),
\label{eq:teacher-loss}
\end{equation}
where $N_r$ is the number of supervised tokens in $r$.
At the answer boundary, the teacher representation is conditioned on the explicit rationale, while the student representation is conditioned on the complete latent trajectory. Let $H_\psi^l(\cdot)$ denote the layer-$l$ hidden representation at the last token of the answer prefix $p$. The corresponding representations are
\begin{equation}
h_{\mathrm{tea}}^{l}=H_\psi^{l}([x;c;p]),\quad
h_{\mathrm{stu}}^{l}=H_\psi^{l}([x;z_{0:T};p]).
\label{eq:boundary-representations}
\end{equation}
We align the teacher and student representations across a set of layers $M$:
\begin{equation}
\mathcal{L}_{\mathrm{KD}}
=
\frac{1}{|M|}\sum_{l\in M}\frac{1}{s_l}
D_{\mathrm{SL1}}\!\left(h_{\mathrm{stu}}^{l},\operatorname{sg}(h_{\mathrm{tea}}^{l})\right),
\label{eq:distillation-loss}
\end{equation}
where $D_{\mathrm{SL1}}$ denotes smooth $\ell_1$ distance, $\operatorname{sg}$ stops gradients through the teacher representation, and $s_l$ normalizes the representation scale at layer $l$.

\subsection{Relative Answer Potential and Subtrajectory Balance}
\label{sec:subtb-objective}
Sampling multiple latent trajectories can uncover solutions missed by a single path, but stochasticity alone does not ensure effective exploration. We introduce a relative answer potential to evaluate intermediate latent states. Subtrajectory Balance then relates potential differences to transition densities across subtrajectories of different lengths.

\paragraph{Relative answer potential.}
For a sampled state $z_t$, we first measure its support for the gold answer $y^\star$ using the average log-likelihood
\begin{equation}
S_\psi(z_t;x,y^\star)
=
\frac{1}{L}
\sum_{s=1}^{L}
\log p_\psi
\left(
y^\star_s\mid x,z_t,y^\star_{<s}
\right).
\label{eq:state-answer-score}
\end{equation}
Answer likelihood reflects support for the target answer from both the latent state and question context. To assess the contribution of $z_t$, we compare it with a replacement state $\widetilde z_t$ drawn from another question's trajectory at the same reasoning step:
\begin{equation}
\Delta_t
=
S_\psi(z_t;x,y^\star)
-
S_\psi(\widetilde z_t;x,y^\star).
\label{eq:answer-score-contrast}
\end{equation}
Both scores are conditioned on the same question $x$ and gold answer $y^\star$, so $\Delta_t$ measures relative answer support between the sampled and replacement states. We center the contrast at the source state, defining $\Phi_t=\Delta_t-\Delta_0$ with $\Phi_0=0$.

Answer likelihood alone does not determine whether the decoded answer is correct. We therefore augment the terminal potential with a correctness reward $R=\mathbb{I}[\hat y_{\mathrm{greedy}}=y^\star]$, where $\hat y_{\mathrm{greedy}}$ is greedily decoded from the terminal state of the sampled trajectory. The resulting balance potential is
\begin{equation}
\Phi'_t
=
\Phi_t+\mathbb{I}[t=T]R.
\label{eq:balance-potential}
\end{equation}
The potential combines intermediate credit with terminal correctness, while answer supervision propagates through the reparameterized latent trajectory.
\paragraph{Subtrajectory Balance.}
We use Subtrajectory Balance to connect changes in the potential $\Phi'_t$ with stochastic transitions along the sampled trajectory. 
To define the balance objective, we introduce a backward policy $P_B^\eta(z_t\mid z_{t+1},x)$ that models reverse transitions using a separate state-conditioned diagonal Gaussian. For a contiguous subtrajectory $\tau_{i:j}=(z_i,\ldots,z_j)$, where $0\leq i<j\leq T$, the potential change across the subtrajectory is related to the accumulated forward and backward densities $P_F^\theta(\tau_{i:j}\mid x)=\prod_{t=i}^{j-1}P_F^\theta(z_{t+1}\mid z_t,x)$ and $P_B^\eta(\tau_{i:j}\mid x)=\prod_{t=i}^{j-1}P_B^\eta(z_t\mid z_{t+1},x)$.
The balance residual for a subtrajectory $\tau_{i:j}$ is defined as:
\begin{equation}
\begin{aligned}
\delta_{i,j}
&=
\Phi'_i-\Phi'_j
+
\frac{1}{d}
\Bigl[
\log P_F^\theta(\tau_{i:j}\mid x)\\
&\hspace{3.5em}
-
\log P_B^\eta(\tau_{i:j}\mid x)
\Bigr].
\end{aligned}
\label{eq:subtrajectory-residual}
\end{equation}
The factor $1/d$ normalizes the log-density ratio by the latent dimension. A zero residual matches the normalized log-density ratio to the potential change $\Phi'_j-\Phi'_i$ over the subtrajectory.

We impose the balance constraint over all contiguous subtrajectories, yielding the multi-scale objective:
\begin{equation}
\mathcal{L}_{\mathrm{SubTB}}
=
\frac{\sum_{0\leq i<j\leq T}\lambda^{j-i}\delta_{i,j}^{2}}
{\sum_{0\leq i<j\leq T}\lambda^{j-i}}.
\label{eq:subtrajectory-loss}
\end{equation}
The length weight $\lambda\in(0,1]$ controls the relative contribution of subtrajectories. Smaller $\lambda$ emphasizes shorter subtrajectories, whereas $\lambda=1$ weights all subtrajectories equally. All subtrajectory residuals include potential differences and transition densities, with an additional terminal correctness reward for those ending at $T$.

\subsection{Training Objective and Inference}
\label{sec:training-inference}
\method{} combines answer and teacher supervision with the Subtrajectory Balance objective to jointly optimize the stochastic latent policy. The full training objective is:
\begin{equation}
\begin{aligned}
\mathcal{L}
&=
\mathcal{L}_{\mathrm{ans}}
+\lambda_{\mathrm{CoT}}\mathcal{L}_{\mathrm{CoT}}
+\lambda_{\mathrm{KD}}\mathcal{L}_{\mathrm{KD}}\\
&\quad
+\lambda_{\mathrm{flow}}\mathcal{L}_{\mathrm{SubTB}}
+\lambda_{\mathrm{drift}}\mathcal{L}_{\mathrm{drift}}.
\end{aligned}
\label{eq:joint-training-objective}
\end{equation}
To stabilize the residual transitions without directly regularizing stochasticity, we penalize the magnitude of the mean updates, $\mathcal{L}_{\mathrm{drift}}=\frac{1}{Td}\sum_{t=0}^{T-1}\|v_\theta(z_t,x)\|_2^2$. The coefficients $\lambda_{\mathrm{CoT}}$, $\lambda_{\mathrm{KD}}$, $\lambda_{\mathrm{flow}}$, and $\lambda_{\mathrm{drift}}$ weight these objectives.
\paragraph{Inference.}
Given a question $x$, we initialize $z_0=f_\psi(x)$ and sample $T$ successive transitions $z_{t+1}\sim P_F^\theta(\cdot\mid z_t,x)$ from the learned forward policy.
The terminal state $z_T$ conditions answer generation. 

\section{Experimental Setup}
\label{sec:experimentssetup}
Our experiments assess reasoning accuracy and OOD generalization, examine how solution coverage scales with latent sampling, and investigate how the relative answer potential and Subtrajectory Balance shape stochastic latent exploration.
\paragraph{Datasets and tasks.}
We consider both arithmetic and deductive reasoning tasks. For arithmetic reasoning, we train on GSM8K-Aug \citep{deng2023implicit}, an augmented version of GSM8K \citep{cobbe2021training}, and evaluate on the GSM8K and four OOD benchmarks: GSM8K-Hard \citep{gao2023pal} with larger numbers, SVAMP \citep{patel2021nlp} with varied linguistic structures, MultiArith \citep{roy2015solving} with multi-step arithmetic, and ASDiv \citep{miao2020diverse} with diverse problem types. For deductive reasoning, we train on ProsQA \citep{hao2024training} and evaluate on its official test set and four OOD splits.
The 7-hop and 8-hop splits increase reasoning depth, while DAG and Dense vary the graph structure. 

\paragraph{Baselines and models.}
We compare \method{} with four representative reasoning baselines. CoT-SFT \citep{wei2022chain} provides an explicit CoT reference, while COCONUT \citep{hao2024training} and CODI \citep{shen2025codi} represent two approaches to continuous reasoning through curriculum training and self-distillation, respectively. We further include CoLaR \citep{tan2026think}, which enables stochastic exploration through a probabilistic latent head. We use CoLaR-5, whose average latent reasoning length is approximately six steps, providing a stochastic baseline with a comparable reasoning budget. Experiments on GSM8K-Aug use GPT-2 Small \citep{radford2019language} and Llama-3.2-1B-Instruct \citep{grattafiori2024llama}, covering both a compact pretrained model and a modern instruction-tuned backbone. For ProsQA, we additionally evaluate DeepSeek-R1-Distill-Qwen-1.5B \citep{guo2025deepseek} to examine the methods on a reasoning-oriented distilled model. Training configurations and evaluation protocols are detailed in Appendix~\ref{app:configuration}.

\begin{table*}[t]
\centering
\footnotesize
\setlength{\tabcolsep}{6pt}
\renewcommand{\arraystretch}{1.05}
\caption{Arithmetic reasoning results on GSM8K and four OOD benchmarks. Accuracy (Acc.\ \%) reports the mean and sample standard deviation over five inference runs. \#L denotes reasoning length, and $^\dagger$ indicates publicly released checkpoints.}
\label{tab:gsm8k}
\vspace{2pt}
\begin{threeparttable}
\begin{tabular}{lcccccccc}
\toprule
& \multicolumn{2}{c}{In-domain}
& \multicolumn{6}{c}{Out-of-domain} \\
\cmidrule(lr){2-3}
\cmidrule(lr){4-9}
Method
& \multicolumn{2}{c}{GSM8K test}
& GSM8K-Hard & SVAMP & MultiArith & ASDiv
& \multicolumn{2}{c}{OOD Avg.} \\
\cmidrule(lr){2-3}
\cmidrule(lr){4-7}
\cmidrule(lr){8-9}
& Acc. & \#L
& Acc. & Acc. & Acc. & Acc.
& Acc. & \#L \\
\midrule

\rowcolor{black!7}
\multicolumn{9}{c}{\textsc{GPT-2 Small}} \\
\addlinespace[2pt]
CoT-SFT
& \pmval{41.70}{0.82} & $27.33$
& \pmval{8.46}{0.14} & \pmval{39.18}{0.60}
& \pmval{84.33}{0.99} & \pmval{49.60}{0.44}
& $45.39$ & $18.84$ \\
COCONUT
& \pmval{30.14}{0.43} & $6.00$
& \pmval{6.14}{0.26} & \pmval{30.48}{0.65}
& \pmval{72.22}{1.04} & \pmval{38.71}{0.42}
& $36.89$ & $6.00$ \\
CODI
& \pmval{29.53}{0.81} & $6.00$
& \pmval{6.53}{0.09} & \pmval{31.09}{0.21}
& \pmval{78.13}{0.49} & \pmval{40.16}{0.37}
& $38.98$ & $6.00$ \\
\addlinespace[2pt]
\hline
\addlinespace[2pt]
CoLaR-5
& \pmval{7.63}{0.41} & $5.97$
& \pmval{1.77}{0.15} & \pmval{20.80}{0.65}
& \pmval{12.78}{0.97} & \pmval{25.96}{0.23}
& $15.33$ & $4.45$ \\
\method{}
& \pmval{28.20}{0.80} & $6.00$
& \pmval{5.61}{0.26} & \pmval{30.06}{0.72}
& \pmval{60.78}{1.96} & \pmval{40.12}{0.56}
& $34.14$ & $6.00$ \\

\midrule
\rowcolor{black!7}
\multicolumn{9}{c}{\textsc{Llama-3.2-1B-Instruct}} \\
\addlinespace[2pt]
CoT-SFT
& \pmval{52.40}{0.58} & $24.70$
& \pmval{12.45}{0.20} & \pmval{61.42}{0.37}
& \pmval{99.22}{0.63} & \pmval{64.21}{0.40}
& $59.32$ & $16.85$ \\
COCONUT$^{\dagger}$
& \pmval{33.24}{0.63} & $6.00$
& \pmval{7.49}{0.42} & \pmval{47.00}{0.72}
& \pmval{83.67}{1.34} & \pmval{57.62}{0.83}
& $48.94$ & $6.00$ \\
CODI
& \pmval{50.07}{0.37} & $6.00$
& \pmval{11.43}{0.27} & \pmval{55.28}{0.73}
& \pmval{97.22}{0.68} & \pmval{64.86}{0.36}
& $57.20$ & $6.00$ \\
\addlinespace[2pt]
\hline
\addlinespace[2pt]
CoLaR-5$^{\dagger}$
& \pmval{25.46}{0.33} & $5.62$
& \pmval{5.47}{0.13} & \pmval{48.84}{0.27}
& \pmval{87.33}{0.84} & \pmval{53.34}{0.13}
& $48.75$ & $3.92$ \\
\method{} 
& \pmval{34.09}{0.27} & $6.00$
& \pmval{7.67}{0.38} & \pmval{49.00}{0.67}
& \pmval{88.33}{0.81} & \pmval{59.86}{0.29}
& $51.22$ & $6.00$ \\
\bottomrule
\end{tabular}
\end{threeparttable}
\par\vspace{10pt}
\centering
\footnotesize
\setlength{\tabcolsep}{8pt}
\caption{Deductive reasoning results on ProsQA and four structural OOD benchmarks. Accuracy is reported as mean $\pm$ sample standard deviation across three training seeds, each averaged over five inference runs. $^\dagger$ indicates a publicly released checkpoint evaluated over five inference runs.}
\label{tab:prosqa}
\vspace{2pt}
\renewcommand{\arraystretch}{1.02}
\begin{threeparttable}
\begin{tabular}{lc@{\hspace{12pt}}ccccc}
\toprule
& \multicolumn{1}{c}{In-domain}
& \multicolumn{5}{c}{Structural OOD} \\
\cmidrule(lr){2-2} \cmidrule(lr){3-7}
Method & ProsQA test & 7-hop & 8-hop & DAG & Dense & Average \\
\midrule
\rowcolor{black!7}
\multicolumn{7}{c}{\textsc{GPT-2 Small}} \\
\addlinespace[2pt]
CoT-SFT
& \pmval{71.93}{3.63}
& \pmval{11.57}{4.01}
& \pmval{10.50}{4.40}
& \pmval{36.40}{7.98}
& \pmval{22.00}{8.91}
& $20.12$ \\
COCONUT 
& \pmval{98.01}{0.12} 
& \pmval{73.28}{5.58}
& \pmval{72.46}{0.71}
& \pmval{58.68}{1.60}
& \pmval{65.65}{2.68}
& $67.52$ \\
CODI$^{\dagger}$
& \pmval{82.48}{1.03}
& \pmval{69.94}{0.73}
& \pmval{70.07}{0.69}
& \pmval{59.29}{0.89}
& \pmval{65.77}{0.88}
& $66.27$ \\
\addlinespace[2pt]
\hline
\addlinespace[2pt]
CoLaR-5
& \pmval{52.35}{2.82}
& \pmval{29.11}{10.63}
& \pmval{28.47}{10.69}
& \pmval{49.41}{4.57}
& \pmval{27.71}{10.99}
& $33.68$ \\
\method{}
& \pmval{88.93}{0.72}
& \pmval{77.03}{1.06}
& \pmval{77.20}{0.55}
& \pmval{50.70}{1.58}
& \pmval{72.45}{1.63}
& $69.34$ \\
\midrule
\rowcolor{black!7}
\multicolumn{7}{c}{\textsc{LLaMA-3.2-1B-Instruct}} \\
\addlinespace[2pt]
CoT-SFT
& \pmval{95.11}{1.41}
& \pmval{10.60}{1.77}
& \pmval{10.72}{1.48}
& \pmval{88.97}{10.61}
& \pmval{36.14}{9.43}
& $36.61$ \\
COCONUT
& \pmval{98.11}{0.34}
& \pmval{53.37}{6.44}
& \pmval{49.62}{6.09}
& \pmval{65.29}{3.17}
& \pmval{51.33}{3.14}
& $54.90$ \\
CODI
& \pmval{97.47}{1.01}
& \pmval{66.27}{1.85}
& \pmval{65.55}{1.24}
& \pmval{60.45}{0.32}
& \pmval{63.71}{2.27}
& $64.00$ \\
\addlinespace[2pt]
\hline
\addlinespace[2pt]
CoLaR-5
& \pmval{90.60}{1.95}
& \pmval{64.51}{5.74}
& \pmval{53.99}{6.46}
& \pmval{74.81}{1.81}
& \pmval{43.42}{4.52}
& $59.18$ \\
\method{} 
& \pmval{99.52}{0.29} 
& \pmval{64.68}{4.52}
& \pmval{54.05}{8.06}
& \pmval{78.62}{5.58}
& \pmval{42.11}{4.46}
& $59.86$ \\
\midrule
\rowcolor{black!7}
\multicolumn{7}{c}{\textsc{DeepSeek-R1-Distill-Qwen-1.5B}} \\
\addlinespace[2pt]
CoT-SFT 
& \pmval{96.56}{0.30}
& \pmval{10.33}{1.60}
& \pmval{12.41}{0.57}
& \pmval{94.46}{0.42}
& \pmval{47.37}{3.72}
& $41.14$ \\
COCONUT 
& \pmval{98.47}{0.58}
& \pmval{54.10}{8.71}
& \pmval{52.27}{10.21}
& \pmval{63.80}{2.03}
& \pmval{59.50}{4.68}
& $57.42$ \\
CODI 
& \pmval{96.51}{0.06} 
& \pmval{69.77}{3.01}
& \pmval{68.02}{4.28}
& \pmval{59.29}{1.16}
& \pmval{63.11}{3.14}
& $65.05$ \\
\addlinespace[2pt]
\hline
\addlinespace[2pt]
CoLaR-5
& \pmval{92.76}{0.21}
& \pmval{62.30}{5.11}
& \pmval{57.13}{4.70}
& \pmval{63.93}{1.59}
& \pmval{54.91}{7.97}
& $59.57$ \\
\method{}
& \pmval{98.61}{0.24}
& \pmval{68.53}{5.10}
& \pmval{64.18}{0.18}
& \pmval{66.58}{1.46}
& \pmval{43.69}{6.70}
& $60.74$ \\
\bottomrule
\end{tabular}
\end{threeparttable}
\end{table*}
\section{Experimental Results}
\label{sec:main-results}
\subsection{Main results}
\label{sec:main-indomain}
Tables~\ref{tab:gsm8k} and~\ref{tab:prosqa} report in-domain and OOD accuracy on arithmetic and deductive reasoning, comparing explicit CoT (CoT-SFT), deterministic latent reasoning (COCONUT and CODI), and stochastic latent reasoning (CoLaR and \method{}) under the same backbone. 

\paragraph{Arithmetic reasoning.}
\method{} consistently outperforms CoLaR across both backbones, achieving 28.20\% versus 7.63\% on GSM8K with GPT-2 and 34.09\% versus 25.46\% with Llama, alongside higher OOD averages. With GPT-2, COCONUT and CODI retain stronger in-domain and OOD performance. With Llama, however, \method{} surpasses COCONUT on GSM8K and all four OOD benchmarks, while CODI achieves the highest accuracy among latent reasoning methods.

\paragraph{Deductive reasoning.}
The advantage is more pronounced on deductive reasoning. With GPT-2, it achieves the highest OOD average (69.34\%), outperforming COCONUT, CODI, and CoLaR, with leading results on 7-hop, 8-hop, and Dense. With Llama and DeepSeek, \method{} achieves the highest in-domain accuracy (99.52\% and 98.61\%), surpassing all baselines. Its OOD averages exceed those of COCONUT and CoLaR on both backbones, while CODI retains higher averages. These results indicate that the learned latent reasoning paths remain effective under changes in reasoning depth and graph structure.
\begin{figure*}[t]
\centering
\includegraphics[width=\textwidth]
{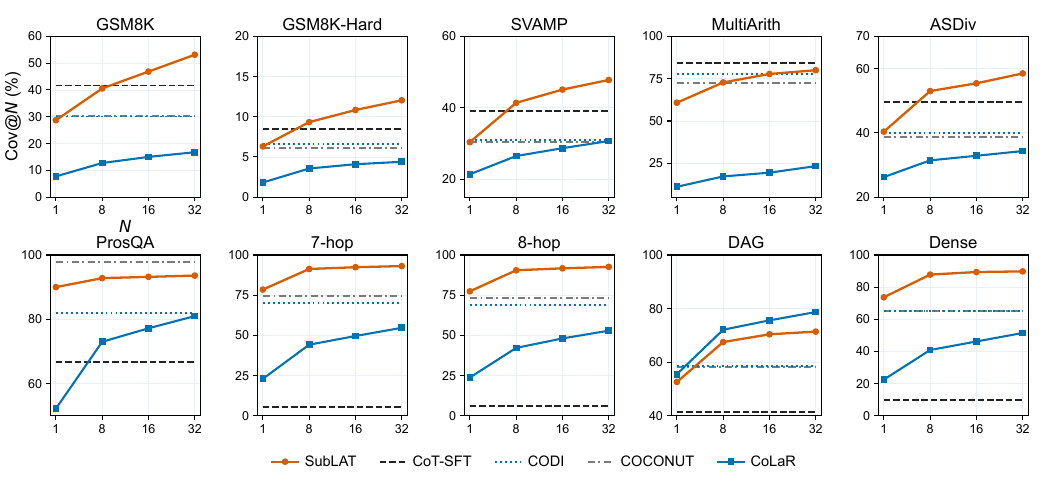}
\caption{Solution coverage (\%) versus sampling budget $N$ on GSM8K (top) and ProsQA (bottom), each with four OOD benchmarks. Orange curves denote \method{}, and blue curves denote CoLaR. Horizontal lines indicate Cov@1 for CoT-SFT, CODI, and COCONUT.}
\label{fig:stochastic-coverage}
\end{figure*}

\subsection{Coverage Scaling with Latent Sampling}
\label{sec:coverage-scaling}
We measure how the coverage scales as the sampling budget increases. COCONUT and CODI reason along a deterministic latent trajectory. For \method{} and CoLaR, we sample a new latent trajectory for each of 32 independent attempts per question. For a sampling budget $N$, solution coverage \citep{you2026parallel} is defined as
\begin{equation}
\operatorname{Cov}@N
=
\frac{1}{|\mathcal D|}
\sum_{x\in\mathcal D}
\mathbbm{1}\!\left[\exists\, k\leq N:\hat y_x^{(k)}=y_x^\star\right],
\label{eq:solution-coverage}
\end{equation}
the fraction of questions answered correctly by at least one of $N$ sampled trajectories, where $\hat y_x^{(k)}$ is the answer from the $k$-th trajectory. 

We evaluate coverage with GPT-2 at sampling budgets $N\in\{1,8,16,32\}$. Figure~\ref{fig:stochastic-coverage} shows that GSM8K coverage increases from 28.47\% at $N=1$ to 53.15\% at $N=32$. \method{} exceeds CoLaR across all evaluated budgets, with a widening gap as $N$ increases. Coverage also increases on all four arithmetic OOD benchmarks, exceeding the accuracy references of COCONUT and CODI at $N=32$.
On ProsQA, \method{} shows consistent coverage growth across the test set and all four structural OOD splits. At $N=32$, coverage reaches approximately 90\% on 7-hop, 8-hop, and Dense, exceeding the references of the deterministic latent baselines on these splits. CoLaR achieves higher coverage on DAG at larger budgets, suggesting that the relative effectiveness of latent sampling varies across graph structures. These results show that solution coverage increases with the sampling budget across arithmetic and deductive reasoning, including under OOD shifts.

\subsection{Latent Exploration and Potential Separation}
\label{sec:trajectory-diversity}
We now examine the structure of these sampled trajectories from two perspectives: how they spread through latent space and how their relative answer potentials evolve with different outcomes.

\paragraph{Latent trajectory diversity.}
Following \citet{you2026parallel},  we measure diversity at step $t$ by the average pairwise cosine dissimilarity among the $K$ sampled latent states of each question,
\begin{equation}
c_t=\frac{1}{|\mathcal D|}
\sum_{x\in\mathcal D}
\frac{2}{K(K-1)}
\sum_{k<l}
\left[1-\cos\!\left(z_t^{(k)},z_t^{(l)}\right)\right].
\label{eq:mean-stepwise-cosine-diversity}
\end{equation}
Here, $z_t^{(k)}$ is the latent state of the $k$-th sampled trajectory at reasoning step $t$.

\begin{figure}[t]
\centering
\includegraphics[width=\columnwidth]
{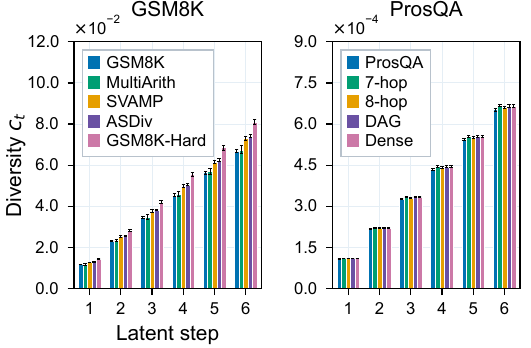}
\caption{Latent trajectory diversity of \method{} across
reasoning steps on GSM8K (left) and ProsQA (right),
including their respective OOD benchmarks.
Error bars indicate 95\% confidence intervals.}
\label{fig:latent-exploration-potential}
\end{figure}

Figure~\ref{fig:latent-exploration-potential} shows that latent trajectory diversity increases steadily across reasoning steps on GSM8K, ProsQA, and their respective OOD benchmarks. Trajectories sampled from the same question remain relatively close at early steps and gradually diverge as latent transitions accumulate. Despite differences in diversity magnitude between arithmetic and deductive reasoning, this pattern holds across both reasoning tasks, indicating that stochastic exploration unfolds progressively over the latent reasoning process.
\paragraph{Potential separation.}
For each question $x$, let $\mu_t^+(x)$ and $\mu_t^-(x)$ denote the mean relative answer potentials $\Phi_t$ of correct and incorrect trajectories at step $t$, respectively. We compute the within-question potential gap over the set $\mathcal M$ of questions with both correct and incorrect trajectories:
\begin{equation}
G(t)=\frac{1}{|\mathcal M|}\sum_{x\in\mathcal M}\left[\mu_t^+(x)-\mu_t^-(x)\right],
\label{eq:mean-potential-gap}
\end{equation}
where a positive $G(t)$ indicates that correct trajectories receive higher potential on average.
\begin{figure}[t]
\centering
\includegraphics[width=0.85\columnwidth]
{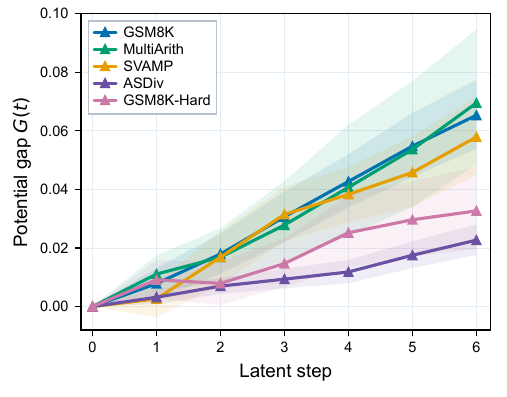}
\caption{Within-question relative answer potential gap $G(t)$ on GSM8K and four arithmetic OOD benchmarks. Gaps are averaged over questions with both outcomes among 32 sampled trajectories per question. Shaded regions indicate 95\% confidence intervals.}
\label{fig:potential-gap}
\end{figure}

Figure~\ref{fig:potential-gap} shows positive mean potential gaps across all five arithmetic benchmarks after the source state, with increasing separation at later reasoning steps. Correct trajectories thus receive higher relative potentials on average, indicating an association between the learned potential and answer correctness.

\subsection{Exploration under Generic Latent Perturbations}
\label{sec:generic-latent-perturbations}
We next examine whether greater latent dispersion alone leads to higher solution coverage. Following \citet{you2026parallel}, we apply MC-dropout and additive Gaussian noise (AGN) to COCONUT and CODI at inference time, varying the dropout rate over $p\in[0.1,0.7]$ and the relative noise scale over $\sigma\in[0.05,0.7]$. 
For \method{}, we instead scale the learned policy standard deviation by $\alpha$, varying the scaling factor over $\alpha\in[1,4]$. We compare coverage against mean pairwise cosine dissimilarity across sampled trajectories, averaged over reasoning steps and questions.

\begin{figure}[t]
\centering
\includegraphics[width=\columnwidth]
{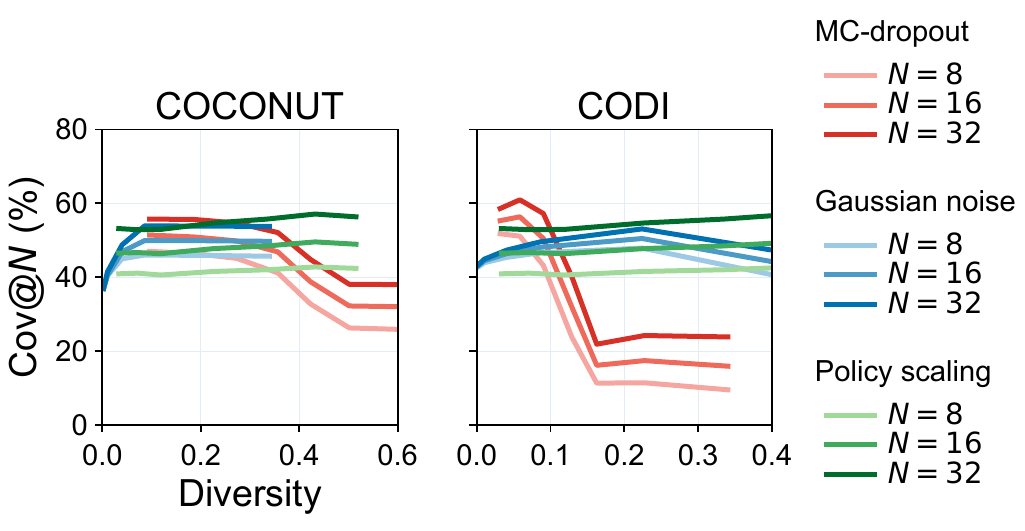}
\caption{Coverage versus latent diversity on GSM8K with GPT-2 at sampling budgets $N$. MC-dropout (red) and AGN (blue) perturb COCONUT (left) and CODI (right), while policy scaling (green) varies the learned transition standard deviation of \method{}.}
\label{fig:diversity-coverage-uplift}
\end{figure}
Figure~\ref{fig:diversity-coverage-uplift} shows that greater latent diversity does not necessarily lead to higher solution coverage. For both COCONUT and CODI, MC-dropout improves coverage at moderate perturbation strengths, but stronger perturbations increase diversity while reducing coverage. AGN yields relatively limited coverage gains as the sampling budget increases. 
Although MC-dropout on CODI reaches a higher peak, its coverage declines under stronger perturbations. Unlike generic perturbations, \method{} achieves strong coverage with its learned stochastic policy, while maintaining coverage across a broader range of exploration scales.

\subsection{Ablation Studies}
\label{sec:ablation}
We examine the contributions of the relative answer potential and Subtrajectory Balance through ablations on GSM8K.
\begin{table}[t]
\caption{Ablations on GSM8K with GPT-2. Greedy uses the mean latent path, Cov@$N$ samples $N$ latent trajectories.}
\centering
\footnotesize
\setlength{\tabcolsep}{4pt}
\begin{tabular}{@{}lcccc@{}}
\toprule
 & & \multicolumn{3}{c}{Cov@$N$ (\%)} \\
\cmidrule(lr){3-5}
Method & Greedy (\%) & $1$ & $8$ & $32$ \\
\midrule
\method{} & 29.71 & 28.47 & 40.56 & 53.15 \\
\quad w/o relative potential
 & 28.32 & 27.03 & 39.20 & 49.76 \\
\quad w/o SubTB
 & 24.94 & 25.32 & 36.38 & 44.03 \\
\bottomrule
\end{tabular}

\label{tab:gsm8k-ablation}
\end{table}
\paragraph{Relative Potential.}
To assess the contribution of relative potential, we replace it with the absolute answer likelihood. This reduces greedy accuracy and coverage across all sampling budgets, indicating that relative potential provides a more informative learning signal for stochastic latent reasoning.

\paragraph{Subtrajectory Balance.}
To assess the contribution of Subtrajectory Balance (SubTB), we remove its training objective. As shown in Table~\ref{tab:gsm8k-ablation}, greedy accuracy drops from 29.71\% to 24.94\%, while coverage decreases across all sampling budgets, reaching 44.03\% at $N=32$ compared to 53.15\% for \method{}. These results indicate that SubTB benefits both greedy reasoning and latent exploration.

\section{Conclusion}
We introduced \method{}, a stochastic latent reasoning framework that combines Subtrajectory Balance with a relative answer potential. Experiments on arithmetic and deductive reasoning demonstrate improved solution coverage with larger sampling budgets. Further analyses show that higher relative potentials are associated with greater answer correctness, while increased latent diversity alone does not guarantee better solution coverage. These findings highlight the importance of effective exploration for reasoning in continuous latent spaces.

\section*{Limitations}
Our experiments focus on arithmetic and deductive reasoning with a fixed latent horizon. Extending \method{} to more complex tasks and adaptive latent horizons remains a direction for future work.
Moreover, the relative answer potential relies on gold answers during training and cannot directly guide answer selection at inference time. While increased sampling improves solution coverage, it does not guarantee correct answer selection. Future work could explore efficient sampling and answer selection strategies to translate coverage gains into reasoning accuracy.

\bibliography{references}

@article{bengio2023gflownet,
  title={Gflownet foundations},
  author={Bengio, Yoshua and Lahlou, Salem and Deleu, Tristan and Hu, Edward J and Tiwari, Mo and Bengio, Emmanuel},
  journal={Journal of Machine Learning Research},
  volume={24},
  number={210},
  pages={1--55},
  year={2023}
}

@inproceedings{lahlou2023theory,
  title={A theory of continuous generative flow networks},
  author={Lahlou, Salem and Deleu, Tristan and Lemos, Pablo and Zhang, Dinghuai and Volokhova, Alexandra and Hern{\'a}ndez-Garc{\i}a, Alex and Ezzine, L{\'e}na N{\'e}hale and Bengio, Yoshua and Malkin, Nikolay},
  booktitle={International Conference on Machine Learning},
  pages={18269--18300},
  year={2023},
  organization={PMLR}
}

@inproceedings{madan2023learning,
  title={Learning gflownets from partial episodes for improved convergence and stability},
  author={Madan, Kanika and Rector-Brooks, Jarrid and Korablyov, Maksym and Bengio, Emmanuel and Jain, Moksh and Nica, Andrei Cristian and Bosc, Tom and Bengio, Yoshua and Malkin, Nikolay},
  booktitle={International Conference on Machine Learning},
  pages={23467--23483},
  year={2023},
  organization={PMLR}
}

@article{whitammer2022trajectory,
  title={Trajectory balance: Improved credit assignment in gflownets},
  author={Whitammer, Esmeralda S and Jain, Moksh and Bengio, Emmanuel and Sun, Chen and Bengio, Yoshua},
  journal={Advances in neural information processing systems},
  volume={35},
  pages={5955--5967},
  year={2022}
}

@inproceedings{shen2025codi,
  title={Codi: Compressing chain-of-thought into continuous space via self-distillation},
  author={Shen, Zhenyi and Yan, Hanqi and Zhang, Linhai and Hu, Zhanghao and Du, Yali and He, Yulan},
  booktitle={Proceedings of the 2025 Conference on Empirical Methods in Natural Language Processing},
  pages={677--693},
  year={2025}
}

@InProceedings{pmlr-v267-yu25k,
  title = 	 {Flow of Reasoning: Training {LLM}s for Divergent Reasoning with Minimal Examples},
  author =       {Yu, Fangxu and Jiang, Lai and Kang, Haoqiang and Hao, Shibo and Qin, Lianhui},
  booktitle = 	 {Proceedings of the 42nd International Conference on Machine Learning},
  pages = 	 {73115--73145},
  year = 	 {2025},
  editor = 	 {Singh, Aarti and Fazel, Maryam and Hsu, Daniel and Lacoste-Julien, Simon and Berkenkamp, Felix and Maharaj, Tegan and Wagstaff, Kiri and Zhu, Jerry},
  volume = 	 {267},
  series = 	 {Proceedings of Machine Learning Research},
  month = 	 {13--19 Jul},
  publisher =    {PMLR},
  url = 	 {https://proceedings.mlr.press/v267/yu25k.html}
}

@misc{you2026slpo,
  title         = {{SLPO}: Scaling Latent Reasoning via a Surrogate Policy},
  author        = {You, Runyang and Liu, Zhiyuan and Li, Yongqi and Li, Wenjie},
  year          = {2026},
  eprint        = {2607.19691},
  archiveprefix = {arXiv},
  primaryclass  = {cs.CL}
}

@misc{zhao2026ltc,
  title         = {Latent Thought Credit: Multi-Answer Credit Assignment for Latent Reasoning},
  author        = {Zhao, Xuyang and Zhang, Liting and Xu, Zichen and Chen, Yong and Zeng, Wenjia and Zhao, Shiwan and Li, Qicheng},
  year          = {2026},
  eprint        = {2608.01593},
  archiveprefix = {arXiv},
  primaryclass  = {cs.CL}
}

@article{williams2026prioritize,
  title={Prioritize the process, not just the outcome: Rewarding latent thought trajectories improves reasoning in looped language models},
  author={Williams, Jonathan and Tureci, Esin},
  journal={arXiv preprint arXiv:2602.10520},
  year={2026}
}

@misc{zou2026ltf,
  title         = {Latent Thought Flow: Efficient Latent Reasoning in Large Language Models},
  author        = {Zou, Xiandong and Huang, Jing and Li, Jianshu and Zhou, Pan},
  year          = {2026},
  eprint        = {2606.16222},
  archiveprefix = {arXiv},
  primaryclass  = {cs.AI},
  url           = {https://arxiv.org/abs/2606.16222}
}

@article{wei2022chain,
  title={Chain-of-thought prompting elicits reasoning in large language models},
  author={Wei, Jason and Wang, Xuezhi and Schuurmans, Dale and Bosma, Maarten and Xia, Fei and Chi, Ed and Le, Quoc V and Zhou, Denny and others},
  journal={Advances in neural information processing systems},
  volume={35},
  pages={24824--24837},
  year={2022}
}

@inproceedings{wang2023selfconsistency,
  title     = {Self-Consistency Improves Chain of Thought Reasoning in Language Models},
  author    = {Wang, Xuezhi and Wei, Jason and Schuurmans, Dale and Le, Quoc V. and Chi, Ed H. and Narang, Sharan and Chowdhery, Aakanksha and Zhou, Denny},
  booktitle = {International Conference on Learning Representations},
  year      = {2023},
  url       = {https://openreview.net/forum?id=1PL1NIMMrw}
}

@inproceedings{xie2026tips,
  title     = {{TIPS}: Turn-Level Information-Potential Reward Shaping for Search-Augmented {LLMs}},
  author    = {Xie, Yutao and Thomas, Nathaniel and Hansen, Nicklas and Fu, Yang and Li, Li Erran and Wang, Xiaolong},
  booktitle = {International Conference on Learning Representations},
  year      = {2026},
  eprint    = {2603.22293},
  archiveprefix = {arXiv},
  url       = {https://arxiv.org/abs/2603.22293}
}

@misc{tao2026trace,
  title         = {{TRACE}: Turn-Level Reward Assignment via Credit Estimation for Long-Horizon Agents},
  author        = {Tao, Leitian and Peng, Baolin and Yao, Wenlin and Ge, Tao and Cheng, Hao and Wang, Mike Hang and Gao, Jianfeng and Li, Sharon},
  year          = {2026},
  eprint        = {2607.13988},
  archiveprefix = {arXiv},
  url           = {https://arxiv.org/abs/2607.13988}
}

@inproceedings{meulemans2023cocoa,
  title     = {Would I Have Gotten That Reward? Long-Term Credit Assignment by Counterfactual Contribution Analysis},
  author    = {Alexander Meulemans and Simon Schug and Seijin Kobayashi and Nathaniel D. Daw and Gregory Wayne},
  booktitle = {Advances in Neural Information Processing Systems},
  volume    = {36},
  year      = {2023},
  doi       = {10.52202/075280-3006},
  url       = {https://proceedings.neurips.cc/paper_files/paper/2023/hash/d8bd445c2abe1343cce0e14b361b2fb3-Abstract-Conference.html}
}

@inproceedings{you2026parallel,
  title={Parallel test-time scaling for latent reasoning models},
  author={You, Runyang and Li, Yongqi and Liu, Meng and Wang, Wenjie and Nie, Liqiang and Li, Wenjie},
  booktitle={Proceedings of the 64th Annual Meeting of the Association for Computational Linguistics (Volume 1: Long Papers)},
  pages={44703--44717},
  year={2026}
}

@article{wang2026gts,
  title={GTS: Inference-Time Scaling of Latent Reasoning with a Learnable Gaussian Thought Sampler},
  author={Wang, Minghan and Bai, Ye and Vu, Thuy-Trang and Shareghi, Ehsan and Haffari, Gholamreza},
  journal={arXiv preprint arXiv:2602.14077},
  year={2026}
}

@inproceedings{zhou2026lepo,
  title={LEPO: Latent Reasoning Policy Optimization for Large Language Models},
  author={Zhou, Yuyan and Yu, Jiarui and Dong, Hande and Hao, Zhezheng and Wang, Hong and Zhang, Jianqing and Lin, Qiang},
  booktitle={Findings of the Association for Computational Linguistics: ACL 2026},
  pages={14416--14427},
  year={2026}
}

@article{deng2023implicit,
  title={Implicit chain of thought reasoning via knowledge distillation},
  author={Deng, Yuntian and Prasad, Kiran and Fernandez, Roland and Smolensky, Paul and Chaudhary, Vishrav and Shieber, Stuart},
  journal={arXiv preprint arXiv:2311.01460},
  year={2023}
}

@article{cobbe2021training,
  title={Training verifiers to solve math word problems},
  author={Cobbe, Karl and Kosaraju, Vineet and Bavarian, Mohammad and Chen, Mark and Jun, Heewoo and Kaiser, Lukasz and Plappert, Matthias and Tworek, Jerry and Hilton, Jacob and Nakano, Reiichiro and others},
  journal={arXiv preprint arXiv:2110.14168},
  year={2021}
}

@inproceedings{gao2023pal,
  title={Pal: Program-aided language models},
  author={Gao, Luyu and Madaan, Aman and Zhou, Shuyan and Alon, Uri and Liu, Pengfei and Yang, Yiming and Callan, Jamie and Neubig, Graham},
  booktitle={International conference on machine learning},
  pages={10764--10799},
  year={2023},
  organization={PMLR}
}

@inproceedings{patel2021nlp,
  title={Are NLP models really able to solve simple math word problems?},
  author={Patel, Arkil and Bhattamishra, Satwik and Goyal, Navin},
  booktitle={Proceedings of the 2021 conference of the North American chapter of the association for computational linguistics: human language technologies},
  pages={2080--2094},
  year={2021}
}

@inproceedings{roy2015solving,
  title={Solving general arithmetic word problems},
  author={Roy, Subhro and Roth, Dan},
  booktitle={Proceedings of the 2015 conference on empirical methods in natural language processing},
  pages={1743--1752},
  year={2015}
}

@inproceedings{miao2020diverse,
  title={A diverse corpus for evaluating and developing English math word problem solvers},
  author={Miao, Shen-Yun and Liang, Chao-Chun and Su, Keh-Yih},
  booktitle={Proceedings of the 58th annual meeting of the Association for Computational Linguistics},
  pages={975--984},
  year={2020}
}

@article{hao2024training,
  title={Training large language models to reason in a continuous latent space},
  author={Hao, Shibo and Sukhbaatar, Sainbayar and Su, DiJia and Li, Xian and Hu, Zhiting and Weston, Jason and Tian, Yuandong},
  journal={arXiv preprint arXiv:2412.06769},
  year={2024}
}

@article{tan2026think,
  title={Think silently, think fast: Dynamic latent compression of llm reasoning chains},
  author={Tan, Wenhui and Li, Jiaze and Ju, Jianzhong and Luo, Zhenbo and Song, Ruihua and Luan, Jian},
  journal={Advances in Neural Information Processing Systems},
  volume={38},
  pages={4646--4668},
  year={2026}
}

@article{grattafiori2024llama,
  title={The llama 3 herd of models},
  author={Grattafiori, Aaron and Dubey, Abhimanyu and Jauhri, Abhinav and Pandey, Abhinav and Kadian, Abhishek and Al-Dahle, Ahmad and Letman, Aiesha and Mathur, Akhil and Schelten, Alan and Vaughan, Alex and others},
  journal={arXiv preprint arXiv:2407.21783},
  year={2024}
}

@article{radford2019language,
  title={Language models are unsupervised multitask learners},
  author={Radford, Alec and Wu, Jeffrey and Child, Rewon and Luan, David and Amodei, Dario and Sutskever, Ilya and others},
  journal={OpenAI blog},
  volume={1},
  number={8},
  pages={9},
  year={2019}
}

@article{guo2025deepseek,
  title={Deepseek-r1: Incentivizing reasoning capability in llms via reinforcement learning},
  author={Guo, Daya and Yang, Dejian and Zhang, Haowei and Song, Junxiao and Wang, Peiyi and Zhu, Qihao and Xu, Runxin and Zhang, Ruoyu and Ma, Shirong and Bi, Xiao and others},
  journal={arXiv preprint arXiv:2501.12948},
  year={2025}
}

@inproceedings{jiang2026rethinking,
  title={Rethinking LLM Reasoning: From Explicit Trajectories to Latent Representations},
  author={Jiang, Cong and Zhang, Xiaofeng and Zhu, Fangzhi and Chen, XiaoWei and Zhu, Junxiong and Zhang, Zheng},
  booktitle={The Fourteenth International Conference on Learning Representations},
  year={2026}
}

@inproceedings{pan2023better,
  title={Better training of gflownets with local credit and incomplete trajectories},
  author={Pan, Ling and Malkin, Nikolay and Zhang, Dinghuai and Bengio, Yoshua},
  booktitle={International Conference on Machine Learning},
  pages={26878--26890},
  year={2023},
  organization={PMLR}
}
\clearpage
\appendix
\section{Experimental Details}
\label{app:configuration}

\subsection{Datasets and Evaluation}
\label{app:datasets}

We train arithmetic reasoning models on GSM8K-Aug, which contains 385{,}620 augmented training examples. Methods requiring reference rationales use 385{,}531 examples after removing records without an available rationale. Checkpoint selection uses a fixed 500-example validation split, with the official GSM8K test set reserved for evaluation. Arithmetic evaluation includes GSM8K (1{,}319 examples), GSM8K-Hard (1{,}319), SVAMP (1{,}000), MultiArith (180), and ASDiv (2{,}249).

ProsQA contains 17{,}886 training, 300 validation, and 500 test examples. We evaluate structural generalization on the 7-hop, 8-hop, DAG, and Dense splits. The 7-hop and 8-hop splits extend reasoning depth, while DAG and Dense introduce changes in graph structure. Table~\ref{tab:dataset-summary} summarizes the datasets and evaluation benchmarks.
\begin{table*}[h]
\centering
\small
\setlength{\tabcolsep}{6pt}
\caption{Training and evaluation benchmarks.}
\label{tab:dataset-summary}
\begin{tabular}{lrrll}
\toprule
Dataset & Training & Validation & In-domain test & Transfer benchmarks \\
\midrule
GSM8K-Aug & 385{,}620 & 500 & GSM8K (1{,}319) & GSM-Hard, SVAMP, MultiArith, ASDiv \\
ProsQA & 17{,}886 & 300 & ProsQA (500)
& 7-hop, 8-hop, DAG, Dense \\
\bottomrule
\end{tabular}
\end{table*}
\subsection{Implementation details.}
We follow the original training procedures for CoT-SFT, COCONUT, CODI, and CoLaR, using the same backbone and training data across methods. COCONUT and \method{} are initialized from the same corresponding CoT-SFT checkpoints. For each setting, the best validation checkpoint is used for test and OOD evaluation. Tables \ref{tab:coconut_hyperparameters}, \ref{tab:codi_hyperparameters}, and \ref{tab:ours-hyperparameters} summarize method-specific training configurations, with additional settings for \method{} in Table~\ref{tab:common-config}. \method{} uses $T=6$ stochastic latent transitions and AdamW with a global batch size of 128. The forward and backward policies use learning rates of $10^{-4}$ and $5\times10^{-5}$, respectively. Dataset- and backbone-specific LoRA learning rates and weight decay are reported in Table \ref{tab:ours-hyperparameters}. Following CODI~\citep{shen2025codi}, we remove the final segment of each reference rationale before teacher supervision while retaining the gold answer. During inference, we follow the CoLaR evaluation protocol with a generation temperature of 1.0 and top-$p$ of 0.9. For GSM8K-Aug, we report mean accuracy and sample standard deviation over five inference runs. For ProsQA, we average five inference runs per training seed and report statistics across three seeds, except GPT-2 CODI, which uses one public checkpoint and reports statistics over five inference runs. For Figure \ref{fig:latent-exploration-potential} and Figure \ref{fig:potential-gap}, pointwise 95\% confidence intervals are estimated using 2,000 bootstrap resamples at the question level.

\begin{table*}[t]
\centering
\begin{threeparttable}
\caption{Training hyperparameters for Coconut models.}
\label{tab:coconut_hyperparameters}
\small
\setlength{\tabcolsep}{4pt}
\renewcommand{\arraystretch}{1.12}
\begin{tabularx}{\textwidth}{@{}l*{4}{C}@{}}
\toprule
& GSM8K & \multicolumn{3}{c}{ProsQA} \\
\cmidrule(lr){2-2}\cmidrule(lr){3-5}
Hyperparameter
& GPT-2 & GPT-2 & LLaMA & DeepSeek \\
\midrule
Latent Tokens Per Stage
& 2 & 1 & 1 & 1 \\
Stage 0 Epochs
& 0 & 5 & 5 & 5 \\
Epochs Per Stage
& 3 & 5 & 5 & 5 \\
Max Latent Stage
& 3 & 6 & 6 & 6 \\
Total Epochs
& 37 & 50 & 20 & 20 \\
Batch Size
& 128 & 128 & 128 & 128 \\
Learning Rate
& $1\times10^{-4}$ & $1\times10^{-4}$
& $5\times10^{-5}$ & $5\times10^{-5}$ \\
Weight Decay
& 0.01 & 0.01 & 0.1 & 0.1 \\
BF16 Precision
& \no & \no & \yes & \yes \\
Reset Optimizer Between Stages
& \yes & \yes & \yes & \yes \\
\bottomrule
\end{tabularx}
\end{threeparttable}
\end{table*}

\begin{table*}[t]
\centering
\begin{threeparttable}
\caption{Training hyperparameters for CODI models.}
\label{tab:codi_hyperparameters}
\small
\setlength{\tabcolsep}{4pt}
\renewcommand{\arraystretch}{1.12}
\begin{tabularx}{\textwidth}{@{}l*{4}{C}@{}}
\toprule
& \multicolumn{2}{c}{GSM8K}
& \multicolumn{2}{c}{ProsQA} \\
\cmidrule(lr){2-3}\cmidrule(lr){4-5}
Hyperparameter
& GPT-2 & LLaMA & LLaMA & DeepSeek \\
\midrule
Latent Loss Weight ($\alpha$)
& 1 & 1 & 1 & 1 \\
CoT Loss Weight ($\beta$)
& 1 & 1 & 1 & 1 \\
Distillation Loss Weight ($\delta$)
& 1 & 20 & 1 & 1 \\
Num Latent Tokens
& 6 & 6 & 6 & 6 \\
Total Epochs
& 40 & 10 & 20 & 20 \\
Batch Size
& 128 & 128 & 128 & 128 \\
Learning Rate
& $3\times10^{-3}$ & $8\times10^{-4}$
& $2\times10^{-4}$ & $2\times10^{-4}$ \\
Weight Decay
& 0.1 & 0.1 & 0.01 & 0.01 \\
BF16 Precision
& \yes & \yes & \yes & \yes \\
Projection Dim
& 768 & 2048 & 2048 & 1536 \\
LoRA Rank
& 128 & 128 & 128 & 128 \\
LoRA Alpha
& 32 & 32 & 32 & 32 \\
\bottomrule
\end{tabularx}
\end{threeparttable}
\end{table*}

\begin{table*}[t]
\centering
\begin{threeparttable}
\caption{Training hyperparameters for \method{}.}
\small
\setlength{\tabcolsep}{4pt}
\renewcommand{\arraystretch}{1.12}
\begin{tabularx}{\textwidth}{@{}l*{5}{C}@{}}
\toprule
& \multicolumn{2}{c}{\textbf{GSM8K-Aug}}
& \multicolumn{3}{c}{\textbf{ProsQA}} \\
\cmidrule(lr){2-3}\cmidrule(lr){4-6}
\textbf{Hyperparameter}
& GPT-2 & LLaMA
& GPT-2 & LLaMA & DeepSeek \\
\midrule
Answer Loss Weight & 1 & 1 & 1 & 1 & 1 \\
CoT Loss Weight ($\lambda_{\mathrm{CoT}}$) & 1 & 1 & 1 & 1 & 1 \\
Distillation Loss Weight ($\lambda_{\mathrm{KD}}$) & 1 & 1 & 1 & 1 & 1 \\
SubTB Loss Weight ($\lambda_{\mathrm{flow}}$)
& 1 & 1 & 1 & 1 & 1 \\
Drift Loss Weight ($\lambda_{\mathrm{drift}}$)
& 0.05 & 0.05 & 0.05 & 0.05 & 0.05 \\
Num Latent Tokens
& 6 & 6 & 6 & 6 & 6 \\
Total Epochs
& 40 & 10 & 50 & 20 & 20 \\
Batch Size
& 128 & 128 & 128 & 128 & 128 \\
LoRA Learning Rate
& $3\times10^{-3}$ & $8\times10^{-4}$ & $10^{-4}$ & $10^{-4}$ & $10^{-4}$ \\
Forward Learning Rate
& $10^{-4}$ & $10^{-4}$ & $10^{-4}$ & $10^{-4}$ & $10^{-4}$ \\
Backward Learning Rate
& $5\times10^{-5}$ & $5\times10^{-5}$
& $5\times10^{-5}$ & $5\times10^{-5}$ & $5\times10^{-5}$ \\
Weight Decay
& 0.1 & 0.1 & 0.01 & 0.01 & 0.01 \\
BF16 Precision
& \yes & \yes & \yes & \yes & \yes \\
LoRA Rank
& 128 & 128 & 128 & 128 & 128 \\
LoRA Alpha
& 32 & 32 & 32 & 32 & 32 \\
\bottomrule
\end{tabularx}
\label{tab:ours-hyperparameters}
\end{threeparttable}
\end{table*}

\begin{table*}[t]
\centering
\small
\caption{\method{} training configuration.}
\label{tab:common-config}
\begin{tabular*}{\textwidth}{@{\extracolsep{\fill}}llcc@{}}
\toprule
Configuration & Value & Configuration & Value \\
\midrule
Latent horizon & 6
& Effective batch size & 128 \\
Trajectories per question & 1
& Precision & BF16 \\
Optimizer & AdamW
& SubTB length weight & 0.9 \\
Training seeds
& \multicolumn{3}{l}{12 (GSM8K-Aug) / 11, 12, 13 (ProsQA)} \\
\bottomrule
\end{tabular*}
\end{table*}

\subsection{Training Pseudocode}
\label{app:training-pseudocode}

Figure~\ref{fig:training_implementation} summarizes the training procedure of \method{}. Each batch samples a latent trajectory, evaluates the relative answer potential, and applies Subtrajectory Balance over contiguous latent segments. Answer supervision, CoT teacher learning, boundary distillation, and drift regularization are optimized jointly with the transition objective. The balance objective uses detached potentials, while answer supervision backpropagates through the reparameterized latent trajectory. 

\section{Additional Analyses}
\label{app:additional-analyses}
\subsection{Training Dynamics and Cost}
\begin{figure*}[t]
    \centering
    \includegraphics[width=\textwidth]{
        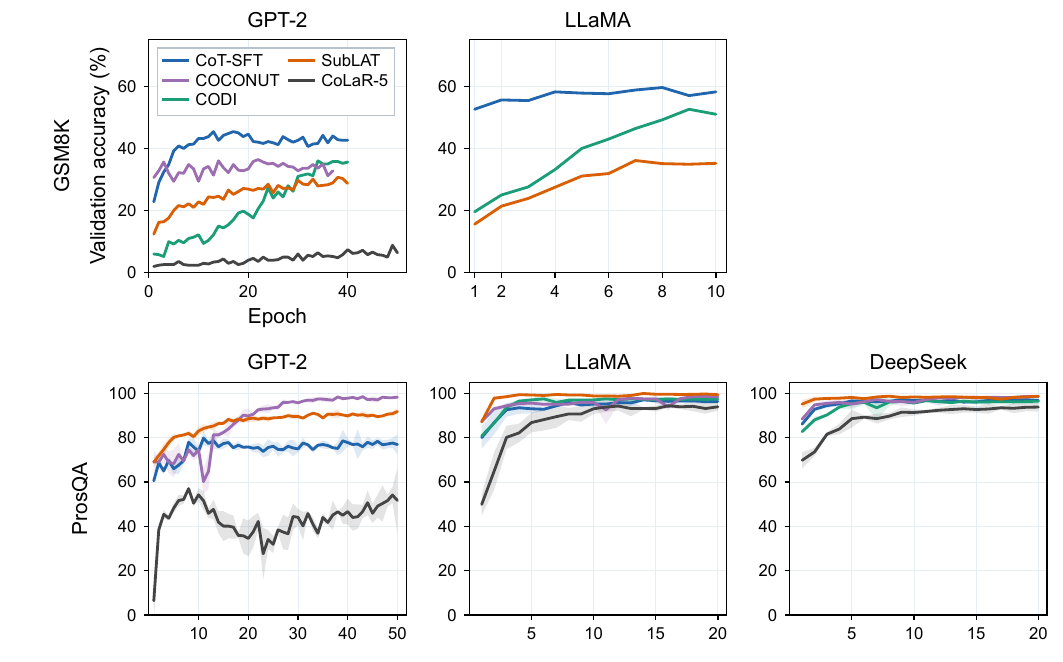}
    \caption{Validation accuracy over training on GSM8K-Aug and ProsQA.}
    \label{fig:training-dynamics}
\end{figure*}

Figure \ref{fig:training-dynamics} shows validation accuracy during training on GSM8K-Aug and ProsQA. Solid lines indicate the mean across available seeds, with shaded regions showing one standard deviation. \method{} improves throughout training on both tasks. On ProsQA, Llama and DeepSeek reach high accuracy early, while GPT-2 converges more gradually, reflecting differences in training dynamics across backbones.

Table \ref{tab:compute-efficiency} reports GPU-hours per epoch and training budgets on GSM8K-Aug. On two H100 GPUs, \method{} requires 6.91 GPU-hours per epoch, compared with 3.71 for COCONUT. The additional cost comes from evaluating sampled and replacement states and optimizing bidirectional transition policies. 
\begin{table*}[t]
\centering
\small
\setlength{\tabcolsep}{3pt}
\caption{Training cost on GSM8K-Aug. GPU-hours per epoch are computed from the number of allocated GPUs and elapsed training time.}
\label{tab:compute-efficiency}
\begin{tabular}{lccc}
\toprule
Method & Hardware & GPU-h/Epoch & Epochs \\
\midrule
CoT-SFT  & 1$\times$GH200 & 0.35 & 40 \\
CODI     & 1$\times$A100  & 0.78 & 40 \\
COCONUT  & 2$\times$H100  & 3.71 & 37 \\
\method{}   & 2$\times$H100  & 6.91 & 40 \\
\bottomrule
\end{tabular}
\end{table*}
\subsection{Qualitative Analysis of Latent States}
\label{sec:latent-decoding}
Table \ref{tab:latent-decoding} presents the top three vocabulary projections at each latent step for COCONUT, CODI, and \method{} on the same GSM8K example, using their respective vocabulary heads. All three methods predict the correct answer, $840$.
The teacher trace includes $30$, which also appears in the projections of COCONUT and CODI. In contrast, \method{} projects $560$ and $280$, corresponding to the partial products in $20\times28+10\times28=840$, with the addition operator appearing near the final step. These projections suggest an alternative decomposition of the problem, distinct from the explicit CoT reference.
This observation is consistent with \method{}'s training design. The relative answer potential and Subtrajectory Balance guide stochastic transitions through answer-based signals, while boundary distillation transfers teacher representations without imposing the explicit CoT sequence on the latent student. 

\begin{figure*}[t]
\centering
\begin{minipage}{0.98\textwidth}
\begin{lstlisting}[
  style=methodcode,
  frame=single,
  basicstyle=\ttfamily\fontsize{7.5pt}{8.5pt}\selectfont\color{black},
  identifierstyle=\color{black},
  keywordstyle=\color{blue!70!black},
  commentstyle=\color{black!75},
  rulecolor=\color{black},
  framesep=5pt,
  aboveskip=0pt,
  belowskip=0pt
]
class CausalSubTBTraining:
    def __init__(self):
        self.num_latent = 6
        self.lambda_subtb = 1
        self.lambda_drift = 0.05
        self.llm = get_gpt2_model()

    def forward(self, x, y, c, p, K=1):
        # x: question; y: answer; c: CoT; p: answer prefix.
        # K: number of sampled latent trajectories per question.

        # Latent rollout and answer CE.
        z, drift = self.rollout(x)
        # Forward and backward log densities averaged over latent dimensions.
        log_pf, log_pb = self.edge_log_probabilities(z, x)
        score = self.probe_trajectory(z, x, y, p)
        answer_loss = -score[:, -1].mean()

        # Counterfactual potential: replace across questions.
        z_cf = (torch.zeros_like(z) if z.shape[0] == K
                else z.detach().roll(-K, dims=0))
        score_cf = self.probe_trajectory(z_cf, x, y, p)
        phi = score - score_cf
        phi = phi - phi[:, :1]

        with torch.no_grad():
            y_hat = self.greedy_decode(x, z[:, -1], p)
            reward = self.answer_exact_match(y_hat, y)
            reward = reward.to(device=phi.device, dtype=phi.dtype)
        
        phi = phi.detach().clone()
        phi[:, -1] += reward.reshape(-1)

        # Subtrajectory balance.
        zero = log_pf.new_zeros(log_pf.shape[0], 1)
        pf = torch.cat([zero, log_pf.cumsum(1)], 1)
        pb = torch.cat([zero, log_pb.cumsum(1)], 1)
        residuals, weights = [], []
        for i in range(z.shape[1] - 1):
            for j in range(i + 1, z.shape[1]):
                residuals.append(phi[:, i] - phi[:, j]
                    + pf[:, j] - pf[:, i] - pb[:, j] + pb[:, i])
                weights.append(0.9 ** (j - i))
        residuals = torch.stack(residuals, 1)
        weights = residuals.new_tensor(weights)
        subtb_loss = ((residuals.square() * weights).sum(1)
                      / weights.sum()).mean()

        # Teacher learning.
        teacher_loss = self._teacher_cot_loss(x, c, y, p)
        ht, valid = self._teacher_boundary_hidden(x, c, p)
        # Student boundary uses the complete latent prefix.
        hs = self._boundary_hidden_from_states(z, x, p)

        # Knowledge distillation.
        layer_losses = []
        if valid.any():
            for student, teacher in zip(hs, ht):
                teacher = teacher.detach().float()
                loss = F.smooth_l1_loss(
                    student.float(), teacher, reduction="none")
                loss = loss.mean(-1)[valid].mean()
                # Normalisation after SmoothL1.
                if valid.sum().item() >= 2:
                    scale = teacher[valid].std().clamp_min(1e-6)
                    scale = torch.nan_to_num(
                        scale, nan=1., posinf=1., neginf=1.)
                    loss = loss / scale
                layer_losses.append(loss)
        kd_loss = (torch.stack(layer_losses).mean()
                   if layer_losses else score.new_zeros(()))

        # Drift regularisation.
        drift_loss = drift.float().square().mean()
        return (answer_loss + teacher_loss + kd_loss
                + self.lambda_subtb * subtb_loss
                + self.lambda_drift * drift_loss)
\end{lstlisting}
\end{minipage}
\caption{Illustrative implementation of \method{} training.}
\label{fig:training_implementation}
\end{figure*}

\begin{table*}[t]
\centering
\caption{Vocabulary projections of latent states on a GSM8K example. Rows 1--3 show the top three tokens at each latent step using each model's vocabulary head, with $\sqcup$ indicating a leading space. Teacher shows CODI's greedy CoT output, and Golden gives the reference solution.}
\label{tab:latent-decoding}

\begingroup
\small
\setlength{\tabcolsep}{3pt}
\renewcommand{\arraystretch}{1.08}
\newcommand{\lvtoken}[1]{\texttt{#1}}
\newcommand{\lvspace}{\ensuremath{\sqcup}}

\begin{tabular*}{\textwidth}
{@{\extracolsep{\fill}}cccccccc@{}}
\toprule
Token rank & 1 & 2 & 3 & 4 & 5 & 6 & Answer \\
\midrule
\multicolumn{8}{c}{COCONUT} \\
\midrule
1
& \lvtoken{\#\#\#}
& \lvtoken{\lvspace30}
& \lvtoken{720}
& \lvtoken{\lvspace30}
& \lvtoken{\lvspace280}
& \lvtoken{\lvspace30}
& \multirow{3}{*}{840} \\
2
& \lvtoken{<|start-latent|>}
& \lvtoken{30}
& \lvtoken{\lvspace packs}
& \lvtoken{\lvspace50}
& \lvtoken{\#\#\#}
& \lvtoken{30}
& \\
3
& \lvtoken{\lvspace with}
& \lvtoken{\lvspace3000}
& \lvtoken{\lvspace get}
& \lvtoken{\lvspace60}
& \lvtoken{\lvspace packs}
& \lvtoken{\lvspace50}
& \\
\midrule

\multicolumn{8}{c}{CODI} \\
\midrule
1
& \lvtoken{+}
& \lvtoken{\lvspace30}
& \lvtoken{s}
& \lvtoken{\lvspace30}
& \lvtoken{s}
& \lvtoken{\lvspace30}
& \multirow{3}{*}{840} \\
2
& \lvtoken{\lvspace of}
& \lvtoken{30}
& \lvtoken{>>}
& \lvtoken{30}
& \lvtoken{>>}
& \lvtoken{30}
& \\
3
& \lvtoken{"}
& \lvtoken{\lvspace40}
& \lvtoken{,}
& \lvtoken{\lvspace40}
& \lvtoken{,}
& \lvtoken{\lvspace40}
& \\
\midrule

\multicolumn{8}{c}{\method{}} \\
\midrule
1
& \lvtoken{40}
& \lvtoken{\lvspace560}
& \lvtoken{\lvspace560}
& \lvtoken{\lvspace560}
& \lvtoken{\lvspace560}
& \lvtoken{\lvspace560}
& \multirow{3}{*}{840} \\
2
& \lvtoken{\lvspace560}
& \lvtoken{40}
& \lvtoken{280}
& \lvtoken{280}
& \lvtoken{280}
& \lvtoken{280}
& \\
3
& \lvtoken{280}
& \lvtoken{280}
& \lvtoken{40}
& \lvtoken{40}
& \lvtoken{40}
& \lvtoken{+}
& \\
\midrule

Teacher
& \multicolumn{6}{l}{\lvtoken{<<20+10=30>>}}
& 840 \\
\addlinespace[2pt]
Golden
& \multicolumn{6}{l}{
  \lvtoken{<<28*20=560>><<10*28=280>><<280+560=840>>}
}
& 840 \\
\bottomrule
\end{tabular*}
\endgroup
\end{table*}

\end{document}